\documentclass[letterpaper, 10 pt, conference]{ieeeconf}  % Comment this line out if you need a4paper

\usepackage{xcolor}
\usepackage[]{hyperref}
\hypersetup{colorlinks,breaklinks,linkcolor=blue,urlcolor=blue,anchorcolor=blue,citecolor=blue}
\usepackage{algorithm,algorithmic}
\usepackage{graphicx}
\usepackage{microtype}
\usepackage{subfigure}
\usepackage{booktabs} % for professional tables
\usepackage{mathtools}
\usepackage{color}
\usepackage{adjustbox}
\usepackage{amssymb}
\usepackage{amsfonts}
\usepackage{multirow}
\usepackage{cite}
\usepackage{bm}
\usepackage[utf8x]{inputenc} 

\IEEEoverridecommandlockouts                              % This command is only needed if 
\title{\LARGE \bf
SPformer: A Transformer Based DRL Decision Making Method for Connected Automated Vehicles
}

\author{
Ye Han, Lijun Zhang$^*$, Dejian Meng, Xingyu Hu, Yixia Lu  % <-this % stops a space
\thanks{Ye Han, Lijun Zhang, Dejian Meng, Xingyu Hu, Yixia Lu are with the School of Automotive Studies, Tongji University, Shanghai 201804, China.
        {\tt\small \{hanye\_leohancnjs, tjedu\_zhanglijun, mengdejian, 2410254, 2051517\}@tongji.edu.cn}}%
\thanks{$^*$Corresponding author: Lijun Zhang}
}

\begin{document}

\maketitle
\thispagestyle{empty}
\pagestyle{empty}

%%%%%%%%%%%%%%%%%%%%%%%%%%%%%%%%%%%%%%%%%%%%%%%%%%%%%%%%%%%%%%%%%%%%%%%%%%%%%%%%
\begin{abstract}
In mixed autonomy traffic environment, every decision made by an autonomous-driving car may have a great impact on the  transportation system. Because of the complex interaction between vehicles, it is challenging to make decisions that can ensure both high traffic efficiency and safety now and futher. Connected automated vehicles (CAVs) have great potential to improve the quality of decision-making in this continuous, highly dynamic and interactive environment because of their stronger sensing and communicating ability. For multi-vehicle collaborative decision-making algorithms based on deep reinforcement learning (DRL), we need to represent the interactions between vehicles to obtain interactive features. The representation in this aspect directly affects the learning efficiency and the quality of the learned policy. To this end, we propose a CAV decision-making architecture based on transformer and reinforcement learning algorithms. A learnable policy token is used as the learning medium of the multi-vehicle joint policy, the states of all vehicles in the area of interest can be adaptively noticed in order to extract interactive features among agents. We also design an intuitive physical positional encodings, the redundant location information of which optimizes the performance of the network. Simulations show that our model can make good use of all the state information of vehicles in traffic scenario, so as to obtain high-quality driving decisions that meet efficiency and safety objectives. The comparison shows that our method significantly improves existing DRL-based multi-vehicle cooperative decision-making algorithms.
\end{abstract}

\section{Introduction}

Autonomus-driving vehicles are playing an increasingly important role in modern transportation systems. At present and for a long time to come, autonomous and human driving vehicles (HDVs) will coexisit both in urban and highway traffic environment. Multi-vehicle collaborative decision-making will play a crucial role in mixed autonomy traffics. It has advantages that single-vehicle autonomous driving cannot match in terms of safety, traffic efficiency, driving experience, energy conservation, and environmental protection\cite{cav-survey-1}. However, due to the dynamic state information and complex interactions of traffic participants, high-quality collaborative driving decision-making is very challenging. Therefore, to develop a good collaborative decision-making algorithm, we should effectively represent the interaction between agents and make full use of it in decision making process.

% Deep reinforcement learning is an effective means to solve multi-agent decision-making problems. It allows agents to gain experience through interaction with the surrounding environment and other traffic participants, and continuously improve their decision-making ability. Connected automated vehicles usually need to weigh between competition and collaboration to meet their own driving purposes and overall traffic efficiency requirements. By reasonably setting reward functions and formulating exploration and exploitation schemes, DRL algorithms can help agents learn high-performance strategies. In addition, , the interaction modeling between agents is also a very important for multi-agent problems.
% 这里有句话没写好

Deep reinforcement learning is an effective method to solve multi-agent decision-making problems. Deep neural networks helps modeling and understanding complex environments and interaction of agent. Reinforcement learning algorithms allow agents to gain experience through interaction with the surrounding environment and other traffic participants, and continuously improve their decision-making ability. Connected automated vehicles usually need to weigh between competition and collaboration to meet their own driving purposes and overall traffic efficiency requirements. By reasonably setting reward functions and formulating exploration and exploitation schemes, DRL algorithms can help agents learn high-performance strategies.

% This paper introduces SPformer, a multi-vehicle collaborative decision-making method based on DRL and transformer architecture. The framework adopts trans\textbf{former} encoders as part of the DRL algorithm. The input of the network is the \textbf{S}tate sequence of all vehicles in the traffic scenario and the output is the multi-vehicle joint driving \textbf{P}olicy. We introduce a learnable policy token as the learning medium of policy and design an intuitive physical positional encoding to improve the algorithm's performance. SPformer can well extract the interactive information between agents, thereby speeding up the learning process of DRL algorithm and improving the quality of learned policy. We verified the algorithm in on-ramp tasks and compared it with the state of the art multi-vehicle decision making algorithms. The results show that our methods have better performance than other deep reinforcement learning algorithms in terms of safety and efficiency.

This paper introduces \textbf{SPformer}, a multi-vehicle collaborative decision-making method based on DRL and transformer architecture. The framework adopts trans\textbf{former} encoders as part of the DRL algorithm. The input of the network is the \textbf{S}tate sequence of all vehicles in the traffic scenario and the output is the multi-vehicle joint driving \textbf{P}olicy. The main contribution of this paper can be summarized as follows:

\begin{itemize}
  \item [1)] 
  An effective multi-vehicle collaborative decision-making framework based on deep reinforcement learning is proposed. It can effectively solve the lateral-longitudinal joint decision making of CAVs from the perspective of mesoscopic traffic.
  \item [2)]
  A learnable policy token are introduced as the learning medium of policy and an intuitive physical positional encoding is designed to improve the algorithm's performance. SPformer can well extract the interactive information between agents, thereby speeding up the learning process of DRL algorithm and improving the quality of learned policy. We verified the algorithm in on-ramp tasks and compared it with the state of the art multi-vehicle decision making algorithms. The results show that our methods have better performance than other deep reinforcement learning algorithms in terms of safety and efficiency.
\end{itemize}

\begin{figure*}[htbp]
\centering
\includegraphics[scale=0.36]{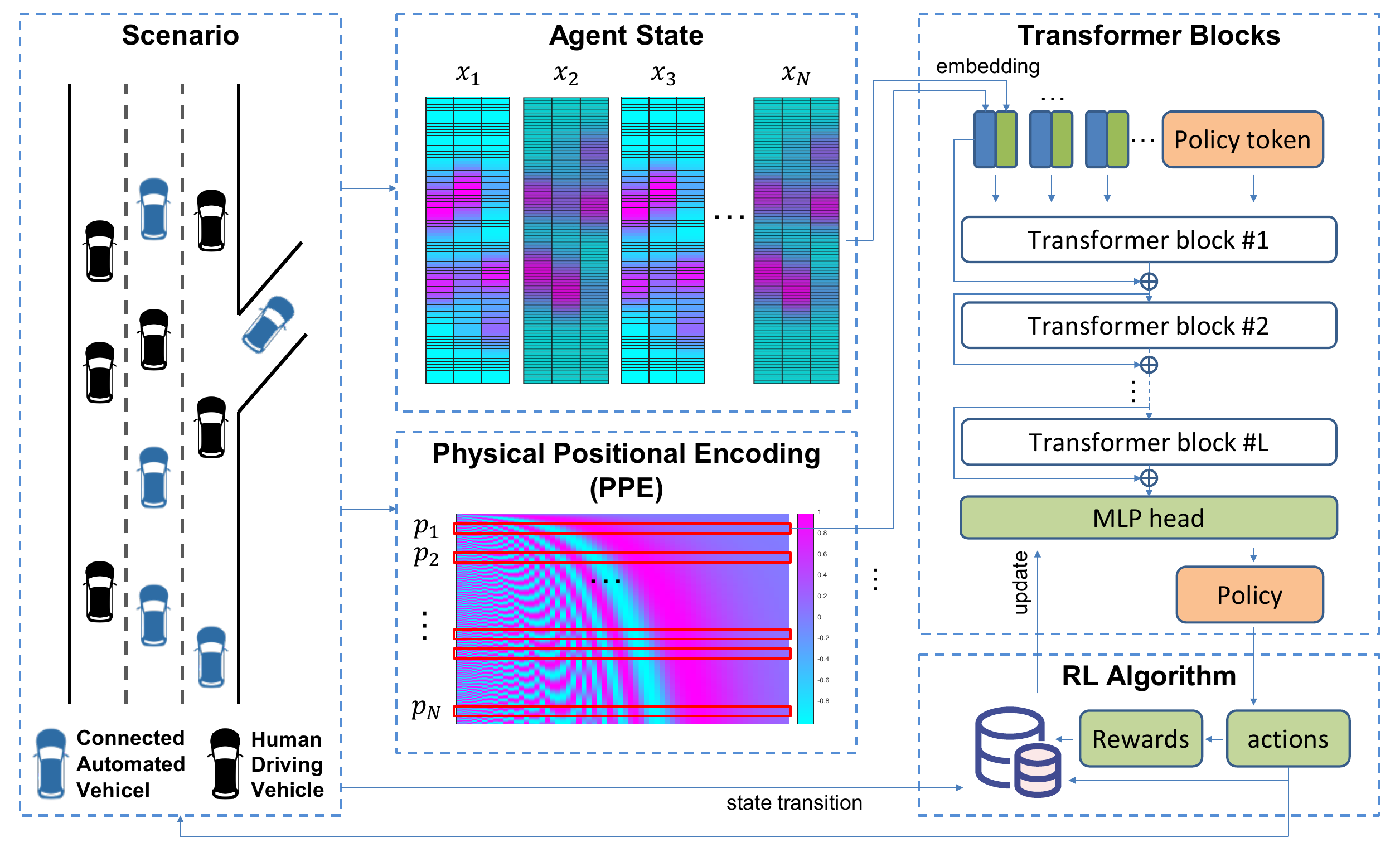}
\caption{An overview of our multi-vehicle decision-making framework with SPformer. Given a mixed autonomy scenario, the vehicle state representation containing multi-modal information is used as the input of SPformer. The positional encoding based on the physical position are added with the embedded vehicle state information, then fed into the transfomer block together with the policy token. The output joint policy could be probability distribution or Q-values of actions, depending on the RL algorithm. The RL algorithm selects actions according to the joint policy and executes them and update the network parameters with collected experience.}
\label{frameworkfig}
\end{figure*}

\section{Related works}

\textbf{Multi-vehicle decision making}:
Multi-vehicle decision-making aims to provide safer and more efficient driving strategies for autonomous driving systems. Early multi-vehicle cooperative decision-making researches can be traced back to the study of longitudinal platooning such as ACC and CACC\cite{cacc-1}. These studies use limited on-board sensors, and the objective is mainly concerned with the string stability in one dimention. Optimization-based planning methods such as mixed integer optimization and dynamic priority allocation can also solve collaborative decision-making problems to some extent\cite{ mixint, dynpri, opt-1}, but it is difficult to guarantee the speed and quality of the solution at the same time in large-scale collaborative driving tasks.

With the development of artificial intelligence, V2X communication, and edge computing technologies, CAVs can make more reasonable decisions in a wider spatial dimension and a longer time range\cite{v2v-1,v2v-2,v2v-3,v2i-1}. The application of deep learning in autonomous driving impels researchers to solve multi-vehicle decision-making problems with DL methods. A. J. M. Muzahid \textit{et al.}\cite{mv-dl-1} systematically summarized the multi-vehicle cooperative collision avoidance technology of CAVs, and proposed a multi-vehicle cooperative perception-communication-decision framework based on deep reinforcement learning. Y. Zheng \textit{et al.}\cite{mv-dl-2} modeled the multi-vehicle decision-making of urban multi-intersections as a predator-pray problem, and used deep reinforcement learning to establish a multi-agent decision-making method where the agents show collaborative behavior patterns far beyond humans. The DL based multi-vehicle decision making algorithms can effectively deal with complex traffic situations, but refined modeling for collaborative interaction is needed for better performance.

% 基于深度学习的多车决策算法具有较好的应用前景，它可以有效应对复杂的交通形势，但是针对多车协同的交互还需要进一步的精细化建模

In addition, game theory, Monte Carlo Tree Search algorithm(MCTS), etc. are also used or combined with deep learning methods to solve multi-vehicle decision-making problems recently\cite{gametheory-2, dec-making-1,gameformer,mcts-1}. These methods have shown great potential in solving problems in complex multi-agent systems.

\textbf{Sequencial interaction modeling}:
Natural language processing (NLP), recommendation systems, time series analysis, etc. all need to properly handle the interaction between sequencial inputs\cite{ interest, nlpsv, timeseries}. RNN and RNN-based LSTM are often used to construct complex sequence interaction models of time series\cite{rnnlstm}. In terms of spatial sequences, A. Alexandre \textit{et al.}\cite{social} proposed social LSTM to predict pedestrian trajectory, and designed a convolutional social pooling to connect the spatial close LSTMs so that information can be shared with each other, which represents the space interaction of agent in complex scenes. Graph neural networks (GNN) introduce graphs to represent the structural relationships between sequences, which are used by many researchers to model the interactions between vehicles in autonomous driving studies. S. Chen \textit{et al.}\cite{ grl-ini } proposed a DRL model combined with GNN to make efficient and safe multi-vehicle cooperative lane change. D. Xu \textit{et al.}\cite{mvgrl-zju} established a multi-vehicle GRL algorithm to realize the cooperative control of vehicles in highway mixed traffic, the graph attention mechanism significantly improves the decision efficiency. In GNN, however, the propagation of information is usually carried out through the adjacency relationship on the graph, which makes long-distance information dissemination difficult.

% 图神经网络的局限性：信息传播有局限性，在GNN中，信息的传播通常是通过图上的邻接关系进行的，特别是长距离传播时效果不佳，并且在位置感知任务中（positionAwareTask）。在某些情况下，长距离的信息传播可能会变得非常困难。 

Transformer is a deep learning architecture with multi-head attention mechanism\cite{transformer}. It has achieved great success in the field of NLP and has been applied to trajectory prediction\cite{ agentformer } and decision making\cite{ qtransformer, dec-transformer } problems. H. Liu \textit{et al.}\cite{ scenerep } implemented two Transformer blocks for scene encoding and vehicle latent feature extraction respectively, which effectively extract the interaction feature between map and agent. The feature is then used by SAC algorithm as input to generate automatic driving policy in different urban driving scenarios. H. Hu \textit{et al.}\cite{ holistic-transformer } finely designed a transformer network to integrate multi-modal information of maps and agents, so as to improve the trajectory prediction and decision-making for autonomous vehicles. Current transformer-based researches on vehicle decision-making use transformer to deal with the multi-modal state sequence input of a single vehicle, and in most cases the sequence is in time order. There are few studies implement transformer architecture in multi-vehicle collaborative decision-making, where the multi-head attention mechanism can properly handle the spatial interaction between agents.

% 当前基于transformer的车辆决策研究多用transformer来处理单个车辆的多模态状态序列输入，对于transformer架构在多车协同决策方面的研究较少，而transformer恰恰可以妥善处理当前。

% In our work, we only use the current state sequence of the agents in the traffic scenario as the input information of the transformer network. And policy-token is introduced as learning medium for multi-agent joint policies. We extend the positional encoding in the basic transformer and proposed an intutive physical positional encoding(PPE). SPFormer can well extract interactive information between vehicles, and has advantages in multi-agent decision-making compared to classic graph neural networks. Simulation tests show that DRL algorithm with SPformer with PPE has faster convergence speed, and the learned driving strategies have better performance in key evaluation metrics.

\section{Problem statement}

This paper aims to solve the collaborative decision-making problem of connected automated vehicles through DRL algorithms. The scenario is mixed traffic where CAVs and HDVs coexist. It is assumed that the CAVs has the ability of global traffic state perception and information sharing in the area of interest. In fact, it is not difficult to realize with the help of roadside facilities and V2X technology. We model the cooperative driving problem from the perspective of mesoscopic traffic flow, considering the lane change and logitudinal acceleration of vehicles, but do not discuss about how these behaviors are realized in terms of vehicle dynamic. This work focuses on the development and verification of interactive collaborative decision-making algorithms between vehicles, and currently does not consider factors such as communication delay and sensing information uncertainty.
\begin{figure}[htbp]
	\vskip -0.1in
	\begin{center}
		\includegraphics[width=\columnwidth]{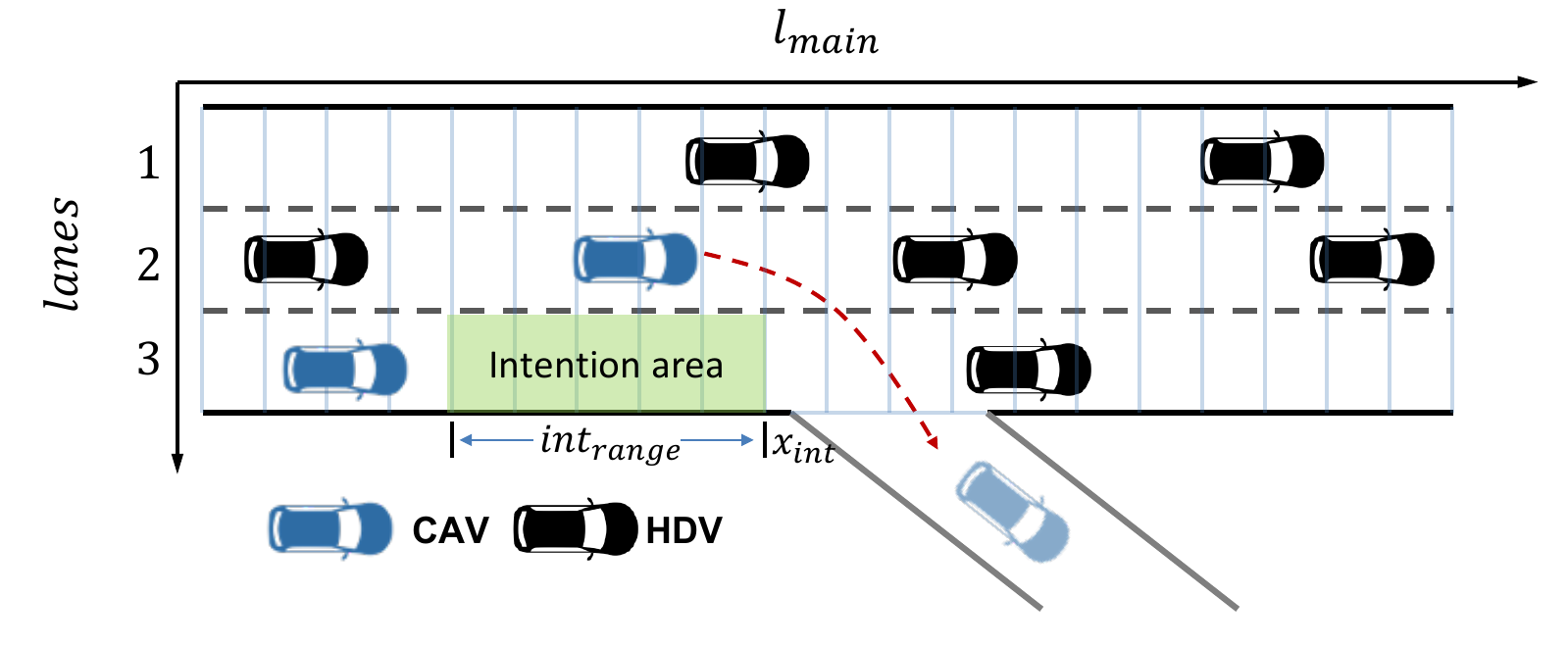}
		\caption{Skatch of off-ramp scene. The main road is rasterized by the lane lines and straight lines equidistant along the center line of the road. For vehicles intending to enter the ramp, the intention area is marked with green.}
		\label{fig-statebase}
	\end{center}
	\vskip -0.1in
\end{figure}

\section{Approach}

\subsection{Deep reinforcement learning problem construction}

% 我们使用多模态耦合的栅格化矩阵来表示车辆的状态
\textbf{State representation}: 
The vehicle's state consists of its individual dynamic characteristics, traffic environment, driving intention, and the relative states of surrounding vehicles. Specifically, vehicle' individual dynamic characteristics include the current lateral and longitudinal position, speed, and acceleration, The traffic environment includes road information, key road element characteristics (locations of intersections, ramps, etc.), The driving intention is the destination of the vehicle in the concerned area. The relative states of surrounding vehicles takes into account all of the other vehicle's relative information to the current vehicle. In this study, we use a multi-modal coupled matrix to represent the state of the vehicle which consists of the informations mentioned above.

The backbone of the state matrix is the rasterized road area.
Take the off-ramp scene shown in Fig.\ref{fig-statebase} as an example. $n_{lanes}$ is the number of main road lanes and $l_{main}$ is the length.
State matrix of the $i$-th vehicle ${\bm{S}_{i} \in \mathbb{R}^{\left(n_{\text {lanes }}+1\right) \times l_{\text {main }}}}$,
and, 

\begin{equation}
\begin{aligned}
\bm{S}_{i} & =\bm{S}_{\text{position},i}+\bm{S}_{\text{speed},i}+\bm{S}_{\text{intention},i}+\bm{S}_{-i} \\
& =\bm{S}_{\text{self},i}+\bm{S}_{-i}
\end{aligned}
\label{statecomp}
\end{equation}
where $\bm{S}_{\text{position},{i}}$, $\bm{S}_{\text{speed},{i}}$, $\bm{S}_{\text{intention},{i}}$, $\bm{S}_{\text{-i}}$ is the ego vehicle's position matrix, velocity field matrix, intention matrix, and relative vehicle information matrix respectively. The sum of the first three is denoted as $\bm{S}_{\text{self},{i}}$.

The position matrix $\bm{S}_{\text {position},{i}}=I_{\text {ego}} \cdot \bm{M}_{\text {position},{i}}$, 
where $\bm{M}_{\text {position},{i}}$ is one hot matrix, where occupied by the ego vehicle is 1, and $I_{ego}$ is the position state factor.

A two-dimensional Gaussian potential field is used for velocity representation. In the above scenario, the speed state matrix of the red vehicle
\begin{equation}
\bm{S}_{\text {speed},{i}}(r, c)=I_{\text {potential}}^{i} v_{i} \cdot e^{-\left[\frac{\left(r-x_{i}\right)^{2}}{2 \sigma_{x}^{2}}+\frac{\left(c-y_{i}\right)^{2}}{2 \sigma_{y}^{2}}\right]}
\end{equation}
where $r$ and $c$ are the row and column index of the velocity state matrix, $I_{\text {potential}}^{i}$ is the speed state factor, $\sigma_{x}$, $\sigma_{y}$ are the longitudinal and lateral speed state decay factors of the vehicle respectively.

A single row matrix $\bm{S}_{\text {intention},{i}}$ is added to represent the location of the target ramp. $\bm{S}_{\text {intention},{i}}$ is initialized as an all-0 matrix and make $x_{\text {int}}^{i}(r, c)=I_{\text {intention}} \text {, if }\left(x_{\text {int}}-\operatorname{int}_{\text {range}}\right)<r<x_{\text {int}} \text { and } c=3$.
where $I_{\text {intention}}$ is the intention state factor, $\text { int }_{\text {range }}$ is the range of the vehicle's intention area.

For vehicle $i$, $\bm{S}_{-i}$ is the weighed sum of the state matrices of all vehicles except itself, i.e.,

\begin{equation}
\bm{S}_{-i}=w_{-}\sum_{j \neq i} \bm{S}_{\text {self},{ j}}
\end{equation}

\begin{figure}[htbp]
\vskip -0.1in
\begin{center}
\includegraphics[width=\columnwidth]{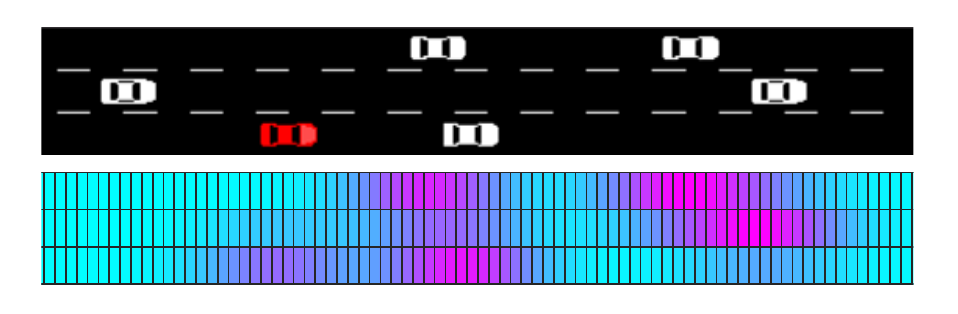}
\caption{An example state representation for a single vehicle. The upper half is the distribution of vehicles on the road section, and the lower half is the state heat map of the red vehicle.}
\label{fig-statemap}
\end{center}
\vskip -0.1in
\end{figure}

So far, the final expression of $\bm{S}_{i}$ is obtained. Fig.\ref{fig-statemap} shows an example state representation for a single vehicle.

\textbf{Action space}: We consider both the lateral and longitudinal behaviors of the vehicle. Longitudinal actions include accelerating, speed keeping and decelerating, and lateral actions consist of left lane changing, lane keeping and right lane changing. Considering that longitudinal and lateral actions can be performed at the same time, there are 9 elements in the joint action space. ${\bm{A}=\left\{\left(a_{\text {lon}}, a_{\text {lat}}\right) | a_{\text {lon}} \in \bm{A}_{\text {lon}}, a_{\text {lat}} \in \bm{A}_{\text {lat}}\right\}}$, where $\bm{A}_{\text {lon}}=\{A C, S K, D C\}$, and $\bm{A}_{\text {lat}}=\{L C, L K, R C\}$.

\textbf{Reward function}: Our work aims at the driving efficiency and safety of the global traffic. The reward function is designed as shown in Equation \ref{eq-rewardfun}, the implementation details can be found in paper\cite{rewardfun}.
\begin{equation}
\begin{aligned}
R&=w_{1} R_{\text {speed }}+w_{2} R_{\text {intention }}+w_{3} P_{\text {collision }}+w_{4} P_{L C} \\
&=\frac{1}{N}(w_{1} {\sum_{i=1}^{N} \frac{v_{i}}{v_{\max }}}+w_{2} N_{\text {onramp }}+w_{3} N_{\text {collision }}+w_{4} N_{L C})
\end{aligned}
\label{eq-rewardfun}
\end{equation}
where $N$ is the number of vehicles in the scene (including HDVs and CAVs), $N_{\text {onramp }}$ is the vehicle passing through intention area at the previous time step and aiming for the ramp, $N_{\text {collision }}$ is the number of collisions, and $N_{L C}$ is the number of frequently lane-changing vehicles.

There are two main differences between our work and the reference work in the calculation of rewards :

\begin{itemize}
  \item [1)] 
  For speed reward, we take into account all vehicles' speed while most previous work only considered the average speed of CAVs. It has been proved that the behavior of autonomous vehicles can affect other vehicles in the traffic environment.      
  \item [2)]
  To encourage vehicles to explore more diverse driving strategies, we only set intention rewards in a small  area close to the ramp as shown in Fig.\ref{fig-statebase}, and no punishment related to intention is set when the vehicles are in other area. 
\end{itemize}

\subsection{Interactive feature extraction method based on Transformer}

In this paper, the transformer encoder is used to extract the interaction features of vehicles. We introduce policy-token as the learning medium of multi-agent joint strategy. The multi-head self-attention mechanism of the transformer helps to extract the interaction information between vehicles. In addition, we integrated physical position encoding into the basic transformer, which makes the network more sensitive to the vehicles’ location and effectively improves the performance of the algorithm.

\textbf{Transformer encoder with policy-token}: 
Inspired by the research in NLP and CV\cite{bert, vit, maptrv2}, we introduce a learnable policy token as the policy learning medium. Policy token enables the network to have perception of global traffic state, it has the same dimension as the vehicle's state feature. $\bm{x}_v \in \mathbb{R}^{H \times W}$ is the input state matrix of each vehicle, it is reshaped to $1\times HW$ and then embedded to $1\times D$. $\bm{x}_{policy} \in \mathbb{R}^{1 \times D}$ is the policy token.

We design the transformer encoder based on ViT\cite{vit}. The encoder consists of multi-head attention layer (MHA) and multi-layer perceptron(MLP) layer alternately. Layernorm(LN) is added before each block, and residual connection is performed after each MHA and MLP. The architecture is shown in the right side of Fig.\ref{frameworkfig} and can be summarized as follows:
\begin{equation}
\begin{aligned}
% F&=H \cdot W \\
\bm{z}_{0}&=\left[\bm{x}_{\text {policy }} ; \bm{x}_{v}^{1} \bm{E} ; \bm{x}_{v}^{2} \bm{E} ; \cdots ; \bm{x}_{v}^{N} \bm{E}\right]+\bm{E}_{\text {pos }} \\
\bm{z}_{\ell}^{\prime}&=\operatorname{MHA}\left(\operatorname{LN}\left(\bm{z}_{\ell-1}\right)\right)+\bm{z}_{\ell-1}, \ell=1 \ldots L \\
\bm{z}_{\ell}&=\operatorname{MLP}\left(\operatorname{LN}\left(\bm{z}_{\ell}^{\prime}\right)\right)+\bm{z}_{\ell}{ }^{\prime}, \ell=1 \ldots L \\
\bm{y}&=\operatorname{LN}\left(\bm{z}_{L}^{0}\right)
\end{aligned} \label{equ_mha}
\end{equation}
where $\bm{E} \in \mathbb{R}^{(H \cdot W) \times D}, \bm{E}_{\text {pos }} \in \mathbb{R}^{(N+1) \times D}$.

We use the standard qkv self-attention to caculate the attention matrix. For the input sequence $\bm{z} \in \mathbb{R}^{N \times D}$, 

\begin{equation}
\begin{aligned}
[\bm{q}, \bm{k}, \bm{v}] &= \bm{z} \cdot \bm{U}_{qkv} \\
\bm{Att}(\bm{z}) &= \operatorname{softmax}\left(\bm{q} \cdot \bm{k}^\top / \sqrt{D_{head}}\right) \cdot \bm{v}
\end{aligned}
\label{eq:qkv}
\end{equation}
where $\bm{U}_{qkv} \in \mathbb{R}^{D \times 3 D_{head}}$, and $\bm{Att}(\bm{z}) \in \mathbb{R}^{N \times N} \label{eq:attn_matrix}$.

The multi-head self-attention is the expansion of self-attention, and $k$ self-attention matrices are calculated simultaneously. The results are concatenated into a multi-head attention matrix.
\begin{equation}
\operatorname{MHA}(\bm{z}) = [\bm{Att}_1(\bm{z}); \bm{Att}_2(\bm{z}); \cdots ; \bm{Att}_k(\bm{z})] \, \bm{W}_{O}
\end{equation}
where $\bm{W}_{O} \in \mathbb{R}^{k \cdot D_{head} \times D}$.

It can be seen that the policy token and the embedded states are fed into the transformer block together, and the final output result is used to derive the policy. The transformer encoder processed a total of $ N+1 $ tokens of dimension $D$, and only the output of the policy token is used to derive the policy. This architecture forces the exchange of information between vehicles' state feature and policy token.

\textbf{Physical positional encoding}:
In NLP studies, word order is of great importantance. Vaswani et al. give the classic position encoding method of sine-cosine alternation. Researchers have improved the PE method according to different tasks, and the performance of transformer has been significantly improved\cite{rethinkpe, 3dppe, peg-vit, complexpe}. For driving tasks, location information is of natural importance. Although location feature is contained in the vehicle states input, it will be diluted by other information such as speed, intention, and maps. To our knowledge, after steps of feature extraction, the superposition of physical location encoding can strengthen location features and improve the network performance.

In this paper, we simply refer to the original transformer, and generate PPE in the form of sines and cosines combination. The map in the area of interest are discretized into $N_{\text {pos}}$ physical positions. For the physical position ${\text{pos}_{ph} - 1}$, the $\bm{PPE}$ is calculated according to Equation \ref{eq-pe}.

\begin{equation}
\begin{aligned}
\bm{PPE}_{2 k}(\cdot, \text { pos}_{ph}) & =\sin \left(\text { pos}_{ph} / (2N_{pos}))^{2 k / D}\right) \\
\bm{PPE}_{2 k+1}(\cdot, \text { pos}_{ph}) & =\cos \left(\text { pos}_{ph} / (2N_{pos}))^{2 k / D}\right)
\end{aligned}
\label{eq-pe}
\end{equation}
where $2k$ and $2k + 1$ are index of PE vector, $D$ is the model dimention.

% 相比于图神经网络，PPE强化了了智能体在环境中的绝对位置信息，我们相信这在场景为中心的任务中能够取得更好的表现。

\begin{figure}[htbp]
\vskip -0.1in
\begin{center}
\includegraphics[width=\columnwidth]{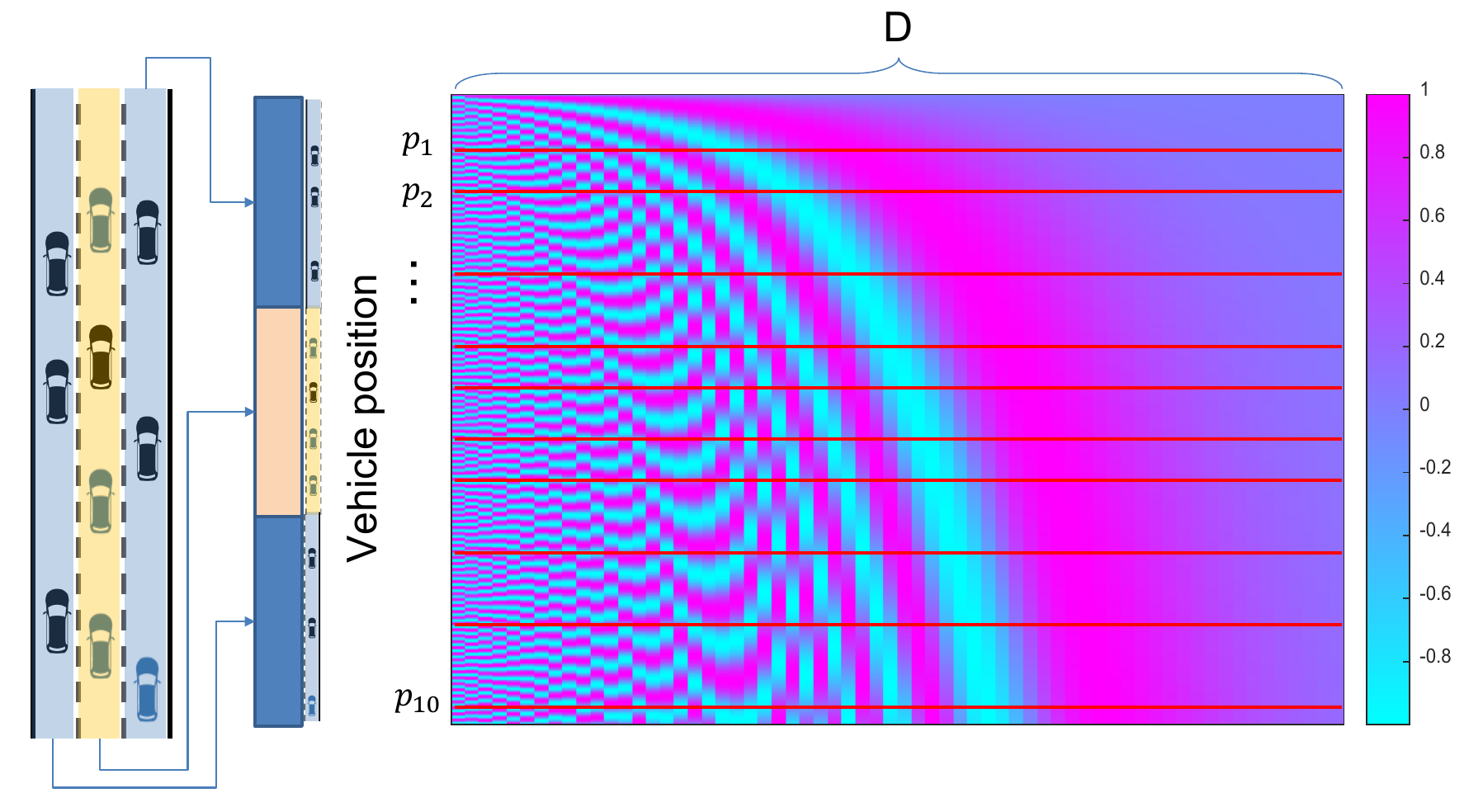}
\caption{Example of PPE. A section of the one-way three-lane road is rejoint by lanes from left to right. The PPE map shown in the right is caculated according to Equation \ref{eq-pe}, the red lines mark the position encoding of all vehicles at this time.}
\label{PPEill}
\end{center}
\vskip -0.1in
\end{figure}

An example of PPE is shown in Fig.\ref{PPEill}. Compared with graph neural networks, PPE enhances the absolute position information of the agent in the environment, which we believe can achieve better performance in scene-centric tasks.

\section{Experiment}

We carried out the verification of the algorithm on the Flow platform\cite{flowframe}. Using DQN as the basic reinforcement learning algorithm, the performance of different deep learning networks is compared.
\begin{table}[h]
	\centering
	% \footnotesize
	\caption{Agent parameters settings for the experiments}
	\begin{adjustbox}{max width=0.45\textwidth}
		\begin{tabular}{ll}
			\hline
			\multicolumn{1}{c}{Parameters} & \multicolumn{1}{c}{Value} \\ \hline
			Number of HDVs       & 4                         \\
			Number of CAVs      & 2                         \\
			HDV departure speed  & 10 $m/s$                      \\
			CAV departure speed & 10 $m/s$                     \\
			Acceleration        & 3.5 $m/s^2$                   \\
			Max HDV speed        & 20 $m/s$                     \\
			Max CAV speed       & 20 $m/s$                     \\
			Initial position	& [20, 30, 50, 50, 30$^*$, 0$^*$]\\
			Initial lane		& [1, 0, 0, 2, 2$^*$, 1$^*$]\\
			Simulation step     & 1 $s$                        \\ \hline
		\end{tabular}
	\end{adjustbox}
	\label{agentparameter}
\end{table}

\subsection{Simulation environment and experiment settings}

We use Flow to build simulation scenarios and verify the algorithm. Flow is a computational framework for deep RL and control experiments for traffic microsimulation. It provides a set of basic traffic control scenarios and tools for designing custom traffic scenarios. In the simulation, the built-in EIDM model of the framework is implemented as HDVs\cite{idm-sumo}. In order to maximize the ability of the algorithm, all active safety detection of the vehicle controlled by the reinforcement learning algorithm are removed during the training process.

The simulation scenario is the on-ramp scenario shown in Fig.\ref{fig-statebase}. Considering a one-way three-lane main road with length of $250 m$, the exit ramp is $200 m$ away from the start. The agents in the case include 2 CAVs and 4 HDVs. 2 CAVs are trageted to enter the ramp, and HDVs are set to drive along the main road until the simulation ends. At the beginning of the simulation, all vehicles are generated at the initial position with the initial speed. Episode ends when the first vehicle in the environment reached the end of main road. Specified parameters of the agent are shown in Table \ref{agentparameter}. The initial position and lane of CAVs are marked by *.

% 其中，2辆自动驾驶车辆的目标为进入匝道，HDVs 则设定为沿着主干道路行驶。仿真开始时，所有车辆同时生成在初始位置，获取初始速度。当环境中首个车辆从主干道路驶出时，仿真结束。

\subsection{RL agent implementation details}

We use the classical deep reinforcement learning algorithm DQN to verify the performance of the proposed method.

DQN is a value-based reinforcement learning algorithm. The Q-Learing algorithm maintains a Q-table, and uses the table to store the return obtained by taking action $a$ under each state $s$, that is, the state-value function $Q(s, a)$. But in many cases, the state space faced by reinforcement learning tasks is continuous, and there are infinite states. In this case, the value function can no longer be stored in the form of tables. To solve this problem, we can use a function $Q ( s, a ; \pmb{\theta} )$ to approximate the action-value $Q ( s, a )$, which is called Value Function Approximation. We use neural networks to generate this function $Q ( s, a ; \pmb{\theta} )$, called Deep Q-network, $\pmb{\theta}$ is a parameter for neural network training. DQN introduces the neural network in deep learning, and uses the neural network to fit the Q table in Q-learning, which solves the problem of \textit{dimension disaster}.

For single-agent DQN, we update the neural network weights $\pmb{\theta}$ by minimizing the loss function:

\begin{equation}
\mathcal{L}\left(s, a \mid \pmb{\theta}\right)=\left(r+\gamma \max _{a} Q\left(s^{\prime}, a^{\prime} \mid \pmb{\theta}\right)-Q\left(s, a \mid \pmb{\theta}\right)\right)^{2} .
\label{eq_lossdqn}
\end{equation}

In our work, we use a single neural network to simultaneously predict the Q values of multiple agents. The MADQN architecture is discussed in \cite{madqn}. Since the reward function is designed to be the mean value of the current state values of all agents, the Q value in this condition should be the discounted sum of the state values of all agents. To this end, we design the following loss function:

\begin{equation}
\begin{array}{c}
\mathcal{L}\left(s, a \mid \pmb{\theta}\right)=\left(r+\gamma \frac{1}{N_{CAV}} \sum_{i=1}^{N_{CAV}} \max _{a_{i}} Q_{i}\left(s^{\prime}, a^{\prime}_{i} \mid \pmb{\theta}\right)\right. \\
\left.-\frac{1}{N_{CAV}} \sum_{i=1}^{N_{CAV}} Q_{i}\left(s, a_{i} \mid \pmb{\theta}\right)\right)^{2}.
\end{array}
\label{eq_desloss}
\end{equation}

SPformer is applied to reinforcement learning agents. The overall structure of the network is described by Formula \ref{equ_mha}, where each MLP contains two fully connected layers with Gaussian error linear unit(GELU) between, and the input is state vector of size $6 \times 1000$. The specific parameters of SPformer are shown in Table \ref{spformerparamerters}. The implementation details of DQN are shown in Table.\ref{dqnparameter}.

\subsection{Compared Methods}

We compare the performance of convolutional neural network, graph neural network and SPformer in DQN algorithm. The convolutional neural network has a kernel with size of $4 \times 4$, followed by a two-layer fully connected network, and rectified linear unit(ReLU) is added after each layer. The implementation details of the graph neural network are shown in paper\cite{grl-itsc}.
\begin{table}[h]
	\centering
	\footnotesize
	\caption{SPformer parameters}
	\begin{adjustbox}{max width=0.45\textwidth}
		\begin{tabular}{lll}
			\hline
			\multicolumn{1}{c}{Variable} & \multicolumn{1}{c}{Parameters} & \multicolumn{1}{c}{Value} \\ \hline
			$E$ &Input dimension & $ 6 \times 1000 $ \\
			% $D$ &Embedding size & 128 \\
			$D$ &Model dimension & 192 \\
			$L$ &Transformer block Layers & 2 \\
			$k$ &Number of heads & 6 \\
			$D_{head}$ &Dimension of head & 32 \\
			- &Dropout rate & 0.1 \\
			- &Output dimention & $1 \times 18$\\ \hline
		\end{tabular}
	\end{adjustbox}
	\label{spformerparamerters}
\end{table}

\begin{table}[h]
	\centering
	\footnotesize
	\caption{DQN parameters}
	\begin{adjustbox}{max width=0.45\textwidth}
		\begin{tabular}{ll}
			\hline
			\multicolumn{1}{c}{Parameters} & \multicolumn{1}{c}{Value} \\ \hline
			Training episodes       & 5000                         \\
			Discount factor      & 1                         \\
			Initial exploration rate      & 1                         \\
			Minimum exploration rate  & 0.01                      \\
			Exploration decay rate & 0.996          \\
			Learning rate        & 0.001               \\
			Batch size        & 16                     \\
			Replay buffer capacity       & 4000        \\
			Intention area range       & 5        \\
			$I_{ego}$       & 30        \\
			$I_{potential}$       & 1        \\
			$\sigma_{x}$, $\sigma_{y}$       & 5, 0.7  \\
			$w_{-}$       & 0.5        \\
			$w_{1}$, $w_{2}$, $w_{3}$, $w_{4}$       & 20, 6, -0.05, -80      \\
			Simulation step     & 1s                   \\ \hline
		\end{tabular}
	\end{adjustbox}
	\label{dqnparameter}
\end{table}

All the methods in the comparison experiment (except EIDM) performed 5000 episodes of training in the on-ramp scenario, and each experiment is conducted 6 times with different random seeds(Two random seeds are assigned to 1).the random action selection of RL algorithm in the exploration process, 2).the random parameters of SUMO built-in car-following and lane-changing controller.). The neural network is trained on a single NVIDIA RTX 3090 GPU using PyTorch and Adam optimizer. The training process of a method in a single scenario takes about 3 hours.

% 两个随机种子分别分配给，1)RL算法在探索过程中的随机动作选取,2)SUMO内置跟车及换道控制器的随机参数。

\subsection{Evaluation Metric}

\textbf{Average traffic score (ATS.)} : 
The superiority of cooperative driving is reflected in the efficiency and safety of global traffic flow, although such a strategy is not always optimal for a single vehicle. Therefore, we use Average Traffic Score (ATS) to evaluate the quality of traffic flow. It is calculated as:

\begin{equation}
ATS=\frac{1}{T} \sum_{t=0}^{T-1} R_{t} .
\label{eq:ats}
\end{equation}

where $T$ is the simulation steps of an episode. 

\textbf{Success rate (Succ.\%)} : Percentage of vehicles successfully entering the ramp in all test cases.

\textbf{Collision number (Coll.)} : The average number of collisions per episode in the test case, which indicates the safety of the strategy.

\textbf{Average velocity (Velo.)} : The mean value of average velocity(in $m/s$) of all vehicles per episode.

\subsection{Results and Comparasion}

\begin{figure}[htbp]
\vskip -0.1in
\begin{center}
\includegraphics[width=\columnwidth]{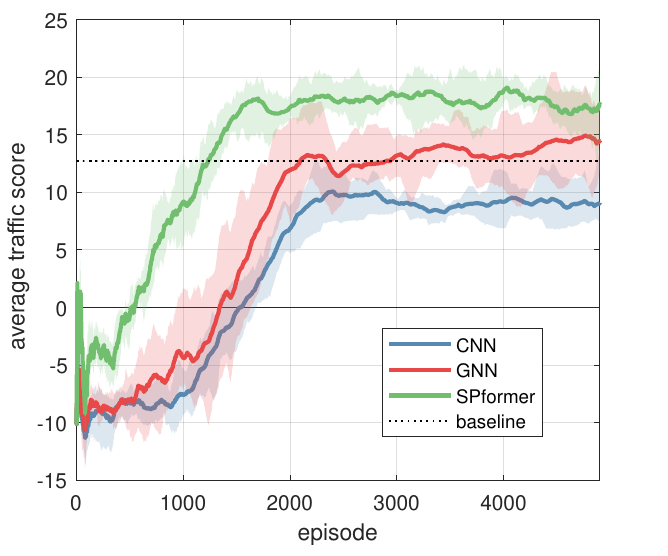}
\caption{Average traffic score in the training process. For each method, we conducted 6 trainings. The curve shown in the figure is the mean ATS value and the shadowed area shows the upper and lower bound of 6 training results.}
\label{resppe}
\end{center}
\vskip -0.0in
\end{figure}

\begin{figure}[h]
	\vskip -0.1in
	\begin{center}
		\includegraphics[width=\columnwidth]{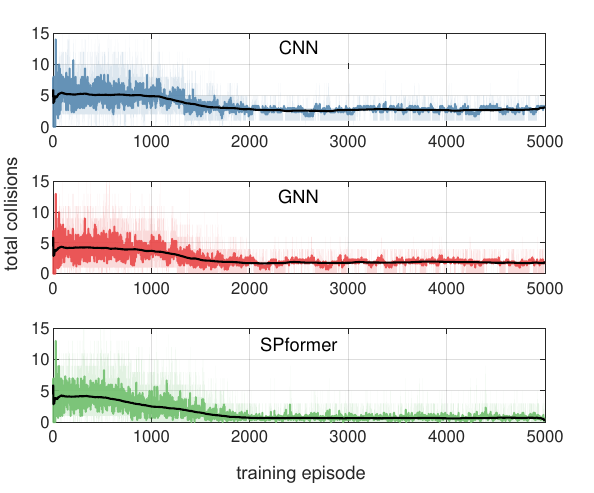}
		\caption{Number of collision in the training process}
		\label{crashfig}
	\end{center}
	\vskip -0.0in
\end{figure}

During the training process, the curves of the average traffic state value and the number of collisions are shown in Fig.\ref{resppe} and Fig.\ref{crashfig}. We use the rule-based approach as a baseline for comparison, which can represent the general level of human drivers.

In the early stage of training, because the agent has a high exploration rate and does not have safety-related experience, the number of collisions is large. This leads to the overall performance of the DRL algorithm agent worse than the rule-based method.

\begin{figure}[h]
	\vskip -0.1in
	\begin{center}
		\includegraphics[width=\columnwidth]{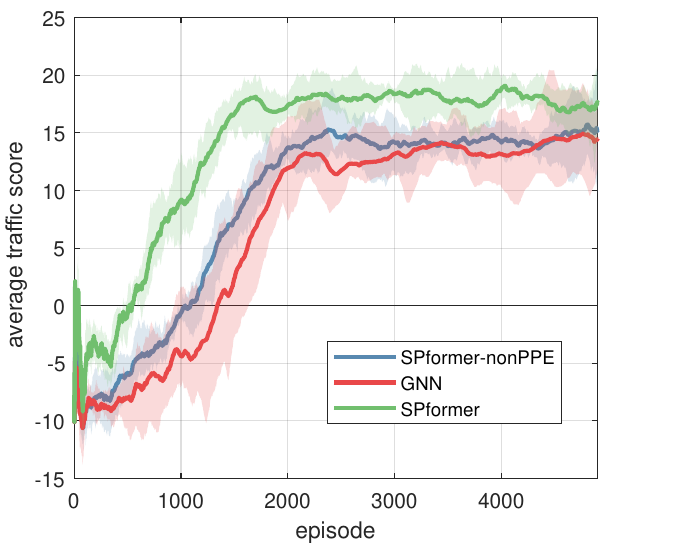}
		\caption{Average traffic score in the training process. This fig shows the importance of PPE in SPformer.}
		\label{resnppe}
	\end{center}
	\vskip -0.0in
\end{figure}

After 2500 episodes of training, the agent has learned a stable strategy. It can be seen in Fig.\ref{resppe} that under the same exploration strategy, the learning efficiency of SPformer is significantly higher than that of CNN and GNN. After 1500 episodes of training, SPformer has already learned a stable driving strategy. This is mainly caused by the additional location information of PPE. It can be seen from the Fig.\ref{resnppe} that after removing PPE, the learning speed of SPformer and the final stable ATS are almost the same as GNN. It should be noted that we have optimized the GNN network to our best with reference to paper\cite{grl-itsc}. The CNN network has also been designed to achieve its best performance in this experiment.

Table \ref{metricscomp} shows that SPformer achieves a good balance between task completion rate, safety and driving speed. Although it does not have an advantage in average speed, it can lead other algorithms significantly on the comprehensive index ATS. This shows that SPformer can fully take into account all agents in the scene and maximize group interests.

\begin{table}[!]
	\centering
	\footnotesize
	\caption{Metrics comparation of 1000 testing episodes}
	\begin{adjustbox}{max width=0.45\textwidth}
		\begin{tabular}{ccccc}
			\hline
			\multicolumn{1}{c}{Method} & \multicolumn{1}{c}{ATS.}& \multicolumn{1}{c}{Succ.\%} & \multicolumn{1}{c}{Coll.} & \multicolumn{1}{c}{Velo.}\\ \hline
			EIDM &12.769 &100 & 0 & 11.200\\
			CNN-DQN &9.134 &62.8 & 2.979 & \pmb{14.764} \\
			GNN-DQN &13.265 &62.8 & 1.548 & 13.958 \\
			SPformer-nonPPE &14.203 &72.6 & 1.407 & 13.931\\
			SPformer &\pmb{18.142} &\pmb{97.4} & \pmb{0.242} & 13.757 \\ \hline
		\end{tabular}
	\end{adjustbox}
	\label{metricscomp}
\end{table}

\section{Conclusion and future work}

In summary, this research proposed SPformer, a DRL-based multi-vehicle collaborative decision-making method, which provides an effective solution to multi-vehicle collaborative lateral and longitudinal joint decision-making problem. SPformer uses policy-token as a learning medium for multi-vehicle driving strategies and integrates an intuitive physical positional encoding. Policy token can prompt the network to obtain a global perception of the traffic state, and physical positional encoding enhances the vehicle location information that is crucial to the quality of decision-making. Therefore, SPformer can effectively improve multi-vehicle cooperative driving strategy learned by DRL algorithms. We tested the performance of SPformer in the on-ramp scenario. Compared with CNN and GNN networks, SPformer have obvious advantages in strategy learning speed and quality.

The future work will focus on improving the performance of cooperative driving algorithms in large-scale scenarios. Although the current algorithm has good interactive decision-making performance, it is difficult to achieve excellent performance in cases with large number of vehicles and random traffic flow, where more diverse cooperative behavior patterns can be found. In addition, physical positional embedding with higher dimension, new architectures combined with game theory and MCTS, and more efficient collaborative state representation methods are to be verified in more complex CAV decision making problems.

% \section*{ACKNOWLEDGMENT}

%%%%%%%%%%%%%%%%%%%%%%%%%%%%%%%%%%%%%%%%%%%%%%%%%%%%%%%%%%%%%%%%%%%%%%%%%%%%%%%%

\bibliographystyle{IEEEtran.bst}
\bibliography{ref.bib}

\begin{thebibliography}{10}
\providecommand{\url}[1]{#1}
\csname url@rmstyle\endcsname
\providecommand{\newblock}{\relax}
\providecommand{\bibinfo}[2]{#2}
\providecommand\BIBentrySTDinterwordspacing{\spaceskip=0pt\relax}
\providecommand\BIBentryALTinterwordstretchfactor{4}
\providecommand\BIBentryALTinterwordspacing{\spaceskip=\fontdimen2\font plus
\BIBentryALTinterwordstretchfactor\fontdimen3\font minus
  \fontdimen4\font\relax}
\providecommand\BIBforeignlanguage[2]{{%
\expandafter\ifx\csname l@#1\endcsname\relax
\typeout{** WARNING: IEEEtran.bst: No hyphenation pattern has been}%
\typeout{** loaded for the language `#1'. Using the pattern for}%
\typeout{** the default language instead.}%
\else
\language=\csname l@#1\endcsname
\fi
#2}}

\bibitem{cav-survey-1}
A.~Talebpour and H.~S. Mahmassani, ``Influence of connected and autonomous
  vehicles on traffic flow stability and throughput,'' \emph{Transportation
  Research Part C: Emerging Technologies}, vol.~71, pp. 143--163, 2016.

\bibitem{cacc-1}
K.~C. Dey, L.~Yan, X.~Wang, Y.~Wang, H.~Shen, M.~Chowdhury, L.~Yu, C.~Qiu, and
  V.~Soundararaj, ``A review of communication, driver characteristics, and
  controls aspects of cooperative adaptive cruise control (cacc),'' \emph{IEEE
  Transactions on Intelligent Transportation Systems}, vol.~17, no.~2, pp.
  491--509, 2015.

\bibitem{mixint}
F.~Fabiani and S.~Grammatico, ``Multi-vehicle automated driving as a
  generalized mixed-integer potential game,'' \emph{IEEE Transactions on
  Intelligent Transportation Systems}, vol.~21, no.~3, pp. 1064--1073, 2020.

\bibitem{dynpri}
S.~Liu, D.~Sun, and C.~Zhu, ``A dynamic priority based path planning for
  cooperation of multiple mobile robots in formation forming,'' \emph{Robotics
  and Computer-Integrated Manufacturing}, vol.~30, no.~6, pp. 589--596, 2014.

\bibitem{opt-1}
Y.~Ouyang, B.~Li, Y.~Zhang, T.~Acarman, Y.~Guo, and T.~Zhang, ``Fast and
  optimal trajectory planning for multiple vehicles in a nonconvex and
  cluttered environment: Benchmarks, methodology, and experiments,'' in
  \emph{2022 International Conference on Robotics and Automation (ICRA)}, 2022,
  pp. 10\,746--10\,752.

\bibitem{v2v-1}
K.~C. Dey, A.~Rayamajhi, M.~Chowdhury, P.~Bhavsar, and J.~Martin,
  ``Vehicle-to-vehicle (v2v) and vehicle-to-infrastructure (v2i) communication
  in a heterogeneous wireless network--performance evaluation,''
  \emph{Transportation Research Part C: Emerging Technologies}, vol.~68, pp.
  168--184, 2016.

\bibitem{v2v-2}
M.~El~Zorkany, A.~Yasser, and A.~I. Galal, ``Vehicle to vehicle “v2v”
  communication: scope, importance, challenges, research directions and
  future,'' \emph{The Open Transportation Journal}, vol.~14, no.~1, 2020.

\bibitem{v2v-3}
H.~Ye, G.~Y. Li, and B.-H.~F. Juang, ``Deep reinforcement learning based
  resource allocation for v2v communications,'' \emph{IEEE Transactions on
  Vehicular Technology}, vol.~68, no.~4, pp. 3163--3173, 2019.

\bibitem{v2i-1}
A.~R. Khan, M.~F. Jamlos, N.~Osman, M.~I. Ishak, F.~Dzaharudin, Y.~K. Yeow, and
  K.~A. Khairi, ``Dsrc technology in vehicle-to-vehicle (v2v) and
  vehicle-to-infrastructure (v2i) iot system for intelligent transportation
  system (its): A review,'' \emph{Recent Trends in Mechatronics Towards
  Industry 4.0: Selected Articles from iM3F 2020, Malaysia}, pp. 97--106, 2022.

\bibitem{mv-dl-1}
A.~J.~M. Muzahid, S.~F. Kamarulzaman, M.~A. Rahman, S.~A. Murad, M.~A.~S.
  Kamal, and A.~H. Alenezi, ``Multiple vehicle cooperation and collision
  avoidance in automated vehicles: Survey and an ai-enabled conceptual
  framework,'' \emph{Scientific reports}, vol.~13, no.~1, p. 603, 2023.

\bibitem{mv-dl-2}
Z.~Yuan, T.~Wu, Q.~Wang, Y.~Yang, L.~Li, and L.~Zhang, ``T3omvp: A
  transformer-based time and team reinforcement learning scheme for
  observation-constrained multi-vehicle pursuit in urban area,''
  \emph{Electronics}, vol.~11, no.~9, 2022.

\bibitem{gametheory-2}
P.~Huang, H.~Ding, Z.~Sun, and H.~Chen, ``A game-based hierarchical model for
  mandatory lane change of autonomous vehicles,'' \emph{IEEE Transactions on
  Intelligent Transportation Systems}, pp. 1--13, 2024.

\bibitem{dec-making-1}
P.~Hang, C.~Huang, Z.~Hu, and C.~Lv, ``Decision making for connected automated
  vehicles at urban intersections considering social and individual benefits,''
  \emph{IEEE Transactions on Intelligent Transportation Systems}, vol.~23,
  no.~11, pp. 22\,549--22\,562, 2022.

\bibitem{gameformer}
Z.~Huang, H.~Liu, and C.~Lv, ``Gameformer: Game-theoretic modeling and learning
  of transformer-based interactive prediction and planning for autonomous
  driving,'' in \emph{Proceedings of the IEEE/CVF International Conference on
  Computer Vision}, 2023, pp. 3903--3913.

\bibitem{mcts-1}
L.~Wen, P.~Cai, D.~Fu, S.~Mao, and Y.~Li, ``Bringing diversity to autonomous
  vehicles: An interpretable multi-vehicle decision-making and planning
  framework,'' \emph{arXiv preprint arXiv:2302.06803}, 2023.

\bibitem{interest}
G.~Zhou, X.~Zhu, C.~Song, Y.~Fan, H.~Zhu, X.~Ma, Y.~Yan, J.~Jin, H.~Li, and
  K.~Gai, ``Deep interest network for click-through rate prediction,'' in
  \emph{Proceedings of the 24th ACM SIGKDD international conference on
  knowledge discovery \& data mining}, 2018, pp. 1059--1068.

\bibitem{nlpsv}
R.~Patil, S.~Boit, V.~Gudivada, and J.~Nandigam, ``A survey of text
  representation and embedding techniques in nlp,'' \emph{IEEE Access},
  vol.~11, pp. 36\,120--36\,146, 2023.

\bibitem{timeseries}
Z.~Mariet and V.~Kuznetsov, ``Foundations of sequence-to-sequence modeling for
  time series,'' in \emph{The 22nd international conference on artificial
  intelligence and statistics}.\hskip 1em plus 0.5em minus 0.4em\relax PMLR,
  2019, pp. 408--417.

\bibitem{rnnlstm}
A.~Sherstinsky, ``Fundamentals of recurrent neural network (rnn) and long
  short-term memory (lstm) network,'' \emph{Physica D: Nonlinear Phenomena},
  vol. 404, p. 132306, 2020.

\bibitem{social}
A.~Alahi, K.~Goel, V.~Ramanathan, A.~Robicquet, L.~Fei-Fei, and S.~Savarese,
  ``Social lstm: Human trajectory prediction in crowded spaces,'' in
  \emph{Proceedings of the IEEE conference on computer vision and pattern
  recognition}, 2016, pp. 961--971.

\bibitem{grl-ini}
S.~Chen, J.~Dong, P.~Y.~J. Ha, Y.~Li, and S.~Labi, ``Graph neural network and
  reinforcement learning for multi-agent cooperative control of connected
  autonomous vehicles,'' \emph{Computer-Aided Civil and Infrastructure
  Engineering}, vol.~36, no.~7, pp. 838--857, 2021.

\bibitem{mvgrl-zju}
D.~Xu, P.~Liu, H.~Li, H.~Guo, Z.~Xie, and Q.~Xuan, ``Multi-view graph
  convolution network reinforcement learning for cavs cooperative control in
  highway mixed traffic,'' \emph{IEEE Transactions on Intelligent Vehicles},
  vol.~9, no.~1, pp. 2588--2599, 2024.

\bibitem{transformer}
A.~Vaswani, N.~Shazeer, N.~Parmar, J.~Uszkoreit, L.~Jones, A.~N. Gomez, L.~u.
  Kaiser, and I.~Polosukhin, ``Attention is all you need,'' in \emph{Advances
  in Neural Information Processing Systems}, I.~Guyon, U.~V. Luxburg,
  S.~Bengio, H.~Wallach, R.~Fergus, S.~Vishwanathan, and R.~Garnett, Eds.,
  vol.~30.\hskip 1em plus 0.5em minus 0.4em\relax Curran Associates, Inc.,
  2017.

\bibitem{agentformer}
Y.~Yuan, X.~Weng, Y.~Ou, and K.~M. Kitani, ``Agentformer: Agent-aware
  transformers for socio-temporal multi-agent forecasting,'' in
  \emph{Proceedings of the IEEE/CVF International Conference on Computer Vision
  (ICCV)}, October 2021, pp. 9813--9823.

\bibitem{qtransformer}
Y.~Chebotar, Q.~Vuong, K.~Hausman, F.~Xia, Y.~Lu, A.~Irpan, A.~Kumar, T.~Yu,
  A.~Herzog, K.~Pertsch, K.~Gopalakrishnan, J.~Ibarz, O.~Nachum, S.~A.
  Sontakke, G.~Salazar, H.~T. Tran, J.~Peralta, C.~Tan, D.~Manjunath, J.~Singh,
  B.~Zitkovich, T.~Jackson, K.~Rao, C.~Finn, and S.~Levine, ``Q-transformer:
  Scalable offline reinforcement learning via autoregressive q-functions,'' in
  \emph{Proceedings of The 7th Conference on Robot Learning}, ser. Proceedings
  of Machine Learning Research, J.~Tan, M.~Toussaint, and K.~Darvish, Eds.,
  vol. 229.\hskip 1em plus 0.5em minus 0.4em\relax PMLR, 06--09 Nov 2023, pp.
  3909--3928.

\bibitem{dec-transformer}
L.~Chen, K.~Lu, A.~Rajeswaran, K.~Lee, A.~Grover, M.~Laskin, P.~Abbeel,
  A.~Srinivas, and I.~Mordatch, ``Decision transformer: Reinforcement learning
  via sequence modeling,'' in \emph{Advances in Neural Information Processing
  Systems}, M.~Ranzato, A.~Beygelzimer, Y.~Dauphin, P.~Liang, and J.~W.
  Vaughan, Eds., vol.~34.\hskip 1em plus 0.5em minus 0.4em\relax Curran
  Associates, Inc., 2021, pp. 15\,084--15\,097.

\bibitem{scenerep}
H.~Liu, Z.~Huang, X.~Mo, and C.~Lv, ``Augmenting reinforcement learning with
  transformer-based scene representation learning for decision-making of
  autonomous driving,'' \emph{IEEE Transactions on Intelligent Vehicles}, pp.
  1--17, 2024.

\bibitem{holistic-transformer}
H.~Hu, Q.~Wang, Z.~Zhang, Z.~Li, and Z.~Gao, ``Holistic transformer: A joint
  neural network for trajectory prediction and decision-making of autonomous
  vehicles,'' \emph{Pattern Recognition}, vol. 141, p. 109592, 2023.

\bibitem{rewardfun}
J.~Dong, S.~Chen, P.~Y.~J. Ha, Y.~Li, and S.~Labi, ``A drl-based multiagent
  cooperative control framework for cav networks: A graphic convolution q
  network,'' \emph{arXiv preprint arXiv:2010.05437}, 2020.

\bibitem{bert}
J.~Devlin, M.-W. Chang, K.~Lee, and K.~Toutanova, ``Bert: Pre-training of deep
  bidirectional transformers for language understanding,'' 2019.

\bibitem{vit}
A.~Dosovitskiy, L.~Beyer, A.~Kolesnikov, D.~Weissenborn, X.~Zhai,
  T.~Unterthiner, M.~Dehghani, M.~Minderer, G.~Heigold, S.~Gelly,
  \emph{et~al.}, ``An image is worth 16x16 words: Transformers for image
  recognition at scale,'' \emph{arXiv preprint arXiv:2010.11929}, 2020.

\bibitem{maptrv2}
B.~Liao, S.~Chen, Y.~Zhang, B.~Jiang, Q.~Zhang, W.~Liu, C.~Huang, and X.~Wang,
  ``Maptrv2: An end-to-end framework for online vectorized hd map
  construction,'' \emph{arXiv preprint arXiv:2308.05736}, 2023.

\bibitem{rethinkpe}
H.~Peng, G.~Li, Y.~Zhao, and Z.~Jin, ``Rethinking positional encoding in tree
  transformer for code representation,'' in \emph{Proceedings of the 2022
  Conference on Empirical Methods in Natural Language Processing}, Y.~Goldberg,
  Z.~Kozareva, and Y.~Zhang, Eds.\hskip 1em plus 0.5em minus 0.4em\relax Abu
  Dhabi, United Arab Emirates: Association for Computational Linguistics, Dec.
  2022, pp. 3204--3214.

\bibitem{3dppe}
C.~Shu, J.~Deng, F.~Yu, and Y.~Liu, ``3dppe: 3d point positional encoding for
  multi-camera 3d object detection transformers,'' \emph{arXiv preprint
  arXiv:2211.14710}, 2022.

\bibitem{peg-vit}
X.~Chu, Z.~Tian, B.~Zhang, X.~Wang, and C.~Shen, ``Conditional positional
  encodings for vision transformers,'' \emph{arXiv preprint arXiv:2102.10882},
  2021.

\bibitem{complexpe}
B.~Wang, D.~Zhao, C.~Lioma, Q.~Li, P.~Zhang, and J.~G. Simonsen, ``Encoding
  word order in complex embeddings,'' \emph{arXiv preprint arXiv:1912.12333},
  2019.

\bibitem{flowframe}
C.~Wu, K.~Parvate, N.~Kheterpal, L.~Dickstein, A.~Mehta, E.~Vinitsky, and A.~M.
  Bayen, ``Framework for control and deep reinforcement learning in traffic,''
  in \emph{2017 IEEE 20th International Conference on Intelligent
  Transportation Systems (ITSC)}.\hskip 1em plus 0.5em minus 0.4em\relax IEEE,
  2017, pp. 1--8.

\bibitem{idm-sumo}
D.~Salles, S.~Kaufmann, and H.-C. Reuss, ``Extending the intelligent driver
  model in sumo and verifying the drive off trajectories with aerial
  measurements,'' \emph{SUMO Conference Proceedings}, vol.~1, p. 1–25, Jul.
  2022.

\bibitem{madqn}
M.~Egorov, ``Multi-agent deep reinforcement learning,'' \emph{CS231n:
  convolutional neural networks for visual recognition}, pp. 1--8, 2016.

\bibitem{grl-itsc}
Q.~Liu, Z.~Li, X.~Li, J.~Wu, and S.~Yuan, ``Graph convolution-based deep
  reinforcement learning for multi-agent decision-making in interactive traffic
  scenarios,'' in \emph{2022 IEEE 25th International Conference on Intelligent
  Transportation Systems (ITSC)}, 2022, pp. 4074--4081.

\end{thebibliography}

\end{document}